\documentclass[12pt]{article}
\usepackage{graphicx}
\usepackage{amsmath}
\usepackage{longtable}
\usepackage{arabtex}
\usepackage{utf8}
\usepackage{lscape}
\usepackage{array}
\usepackage{longtable} 
\usepackage{etoolbox}
\usepackage{url}
\usepackage{fancyvrb}
\usepackage{caption}
\usepackage{color}
\begin{document}

\title{ARCOQ: Arabic Closest Opposite Questions Dataset}

\author{
Sandra Rizkallah,  Amir F. Atiya, and Samir Shaheen\\
Computer Engineering Department\\
Faculty of Engineering, Cairo University\\
Giza, Egypt\\
}

\maketitle
\begin{abstract}
This paper presents a dataset for closest opposite questions in Arabic language. 
The dataset is the first of its kind for the Arabic language. It is beneficial for the assessment of systems on the aspect of antonymy detection. 
The structure is similar to that of the Graduate Record Examination (GRE) closest opposite questions dataset for the English language.
The introduced dataset consists of 500 questions, each contains a query word for which the closest opposite needs to be determined from among a set of candidate words.
Each question is also associated with the correct answer. We publish the dataset publicly in addition to providing standard splits of the dataset into development and test sets.
Moreover, the paper provides a benchmark for the performance of different Arabic word embedding models on the introduced dataset.
\end{abstract} 


﻿\setcode{utf8}
\section{Introduction}
\label{intro}
Antonymy detection is vital for widespread applications including sentiment analysis, contradiction detection, question answering, machine translation and information retrieval \cite{Scheible13,Silveira13,Li17,Bui20}.
{In sentiment analysis, one needs to differentiate between positive and negative attitudes. Some methods were developed for this purpose making use of antoyms and antonymy detection.
In order to assess them, it would be a good idea to have a benchmark dataset for antonyms. This would be helpful in comparing different antonymy detection methods. It can also be beneficial in assessing the capability of the methods in selecting the most accurate antonym.
Educational applications are another example. As a step to automatic answer generation, we need to be able to have an approach that can accurately select the correct answer from a set of choices in multiple choice questions.}

Assessing the capability of a system to detect antonyms is addressed in the English language, for example in \cite{Mohammad08,Mohammad13} a dataset is gathered from Graduate Record Examination (GRE) closest opposite questions. The mentioned dataset is used by a number of works addressing the English language. In \cite{Chang13}, Multi-Relational Latent Semantic Analysis (MRLSA) is presented showing the performance of the model on identifying antonyms relations. In \cite{Zhang14}, a Bayesian-based model is proposed where neural word embeddings are generated to differentiate synonyms and antonyms. In \cite{Ono15}, word embeddings are trained in a way such that antonyms can be captured. In \cite{Jones16}, algorithms are evaluated for detecting antonymy. In \cite{Vulic18}, vector space is fune-tuned to include antonymy relations. In {\cite{Rizkallah20,rizkallah2021new}}, new spherical word embeddings are learned to capture antonymy. In \cite{OMRizkallah20}, an opinion mining approach is presented utilizing antonymy relations. In \cite{Lei21}, contextual relations are addressed with a focus on antonymy.

On the other hand, few works address antonymy in the Arabic language. In \cite{AlHedayani16}, the study investigates antonymy use in the Modern Standard Arabic (MSA) texts. In \cite{AlYahyal16}, ontological lexicon enrichment is performed through semi-automated extraction of lexical relations focusing on antonyms. In \cite{Batita17}, a pattern-based approach is used with the aim of extending the Arabic WordNet with more relations specially, antonyms. In \cite{Rizkallah21}, Arabic word vectors are embedded in a unit sphere to enable antonymy detection where antonyms vectors lie on opposite poles of the sphere. In all the aforementioned works, no unified benchmark is used to measure the performance of the proposed systems.

To our knowledge, datasets that allow for measuring the performance of antonymy detection do not exist for the Arabic language. To fill this gap, we present a new dataset for the Arabic language that can be used for assessing the performance of systems on the aspect of antonymy detection. The dataset contains 500 questions in the form of multiple choice questions. The target is to choose the closest opposite word (antonym) to a query word given in the question from among the given candidate choices. Each question is structured as follows: ``query\_word: candidate\_answer1 candidate\_answer2 candidate\_answer3 :: correct\_answer". Moreover, we present a benchmark that records the performance of different Arabic word embedding models in detecting antonyms on the introduced dataset. {We considered word embedding methods because these are the dominant
approach in antonymy detection.} The dataset is published publicly {\footnote{\url{https://github.com/SandraRizkallah/ARCOQ-dataset}}} in order to urge the research community to build upon the introduced benchmark and create new benchmarks for the Arabic language. 

The rest of the paper is structured as follows: Section 2 discusses the data collection and processing procedures. In Section 3, the benchmark is presented, explaining the methods used and discussing the obtained results on the introduced dataset. Finally, Section 4 shows the conclusion and future work.

\section{Data Collection and Processing}

We collected the dataset from the following sources:
\begin{enumerate}
\item Educational exercises\\
Choosing the closest opposite for a given Arabic word from among a set of candidate words is a question that is commonly included in the educational exercises and exams for the Arabic language subject. We have explored examples of the Arabic subject exams and exercises for the different Egyptian educational stages: primary, preparatory and secondary. Such exercises are provided by the Egyptian Ministry of Education and Technical Education through Egyptian Knowledge Bank (EKB) platform \cite{ekb}.
\item Translating GRE closest opposite questions\\
In \cite{Mohammad08,Mohammad13} a GRE closest opposite questions English dataset is introduced. We have automatically translated all the questions in this dataset from English to Arabic using ``google-trans-new" python package \cite{gt}. {After automatic translation, we manually removed the questions that we considered to be not perfectly translated.}
\item Native Arabic speakers\\
We asked native Arabic  speakers (specifically 4 speakers) to generate questions of the required type and form i.e. one query Arabic word with three candidate words from which the closest opposite of the query word should be chosen. 
\end{enumerate}

After collecting the questions as explained above, we perform some processing and checks:
\begin{itemize}
\item Remove any diacritics (tashkeel).
\item Remove any elongations in the letters (tatweel).
\item Remove any duplicated questions.
\item Make sure that the correct answer is among the given choices in the question (this is a sanity check).
\end{itemize} 

Finally, the dataset consists of 500 questions, each in the following form:\\\\
\begingroup
\fontsize{10}{12}\selectfont ``query\_word: candidate\_answer1 candidate\_answer2 candidate\_answer3 :: correct\_answer"\\
\endgroup
\\
These questions are further sorted alphabetically based on the query word. 
Moreover, we provide standard splits for the dataset into development and test sets. 
We randomly split the dataset into 20\% development set and 80\% test set: 
\begin{itemize}
\item Dev 100: contains 100 questions
\item Test 400: contains 400 questions
\end{itemize} 
{We followed the lead of the GRE dataset \cite{Mohammad08,Mohammad13} in having a development set to test set ratio of 1:4. The reason is that there are not many parameters to tune, for example in most methods pre-trianed embedding vectors are available.}

\begin{figure*}
\centering
\captionsetup{justification=centering}
  \includegraphics[width=0.5\linewidth]{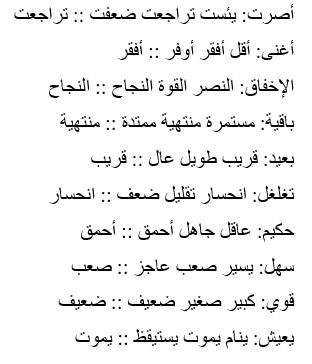}
  \caption{Example Questions from ARCOQ dataset}
  \label{dsex}
\end{figure*}

Figure \ref{dsex} shows example questions from the ARCOQ dataset we established.

\section{Benchmark}

We established a benchmark using our ARCOQ dataset. The performance of different Arabic word embedding models in detecting antonymy is measured. The benchmark includes results for the development set, the test set and the whole dataset. The metrics used for the performance evaluation are those introduced in \cite{Mohammad08,Mohammad13} as follows:
\begin{equation}
Precision=\frac{ \mbox{number of correctly answered questions}}{\mbox{number of answered questions}}
\end{equation}
\begin{equation}
Recall=\frac{ \mbox{number of correctly answered questions}}{ \mbox{total number of questions}}
\end{equation}
\begin{equation}
Fscore= \frac{(2*Precision*Recall)}{(Precision+Recall)}
\end{equation}
A question is answered by the model only if the model has vectors for all the words in the question (query and all candidate answers).

The benchmark includes the major Arabic word embedding models:\\
\begin{enumerate}
\item Polyglot (2013) \cite{Al-Rfou}
\item Altowayan (2016) \cite{Altowayan}
\item AraVec: full grams CBOW 100 twitter (2017) \cite{Soliman}
\item AraVec: full grams SG 100 twitter (2017) \cite{Soliman}
\item fastText (2018) \cite{Grave}
\item ArWordVec: SG\_300\_3 (2020) \cite{Fouad}
\item ArWordVec: SG\_300\_5 (2020) \cite{Fouad}
\item AraBERT (2020) \cite{Antoun}
\item ArSphere (2021) \cite{Rizkallah21}
\end{enumerate}

The models from 1--8 are distributional models that are pre-trained on large Arabic text corpora. In such models, antonyms vectors can be placed far away from each other (small similarity between vectors) or close to each other (high similarity between vectors). The latter is because antonyms often occur in the same context. On the other hand, Arsphere is a semi-supervised model that embeds antonyms vectors at opposite poles of the sphere (negative similarity between vectors).

For answering a given question in the dataset, the following procedure is followed:
\begin{enumerate}
\item Get the vector of the query word from the designated model.
\item Get the vector of each candidate answer from the designated model.
\item Get the similarity scores between the query word and each candidate answer:
\begin{itemize}
\item Normalize all the obtained vectors.
\item Compute the dot product between the normalized vector of the query word and the normalized vector of each candidate answer.
\item Having 3 computed similarity scores, choose the answer of the question as the word with highest/lowest similarity score with the query word.
\item Highest or lowest similarity score?, we performed two experiments:
\begin{itemize}
\item Experiment 1: Lowest similarity score is chosen for all the models.
{Such choice is made since in embedding models the similarity between vectors reflect the underlying semantics so synonyms will have high similarity scores while antonyms or unrelated words conversely have low similarity scores.}
\item Experiment 2: Highest similarity score is chosen for models 1--8 while also the lowest similarity score is chosen for ArSphere.
{This choice is made because in distributional models antonyms often occur in the same context. Therefore, vectors of synonyms as well as antonyms will mostly have high similarity scores.}
\end{itemize}
{The choice for ArSphere model is always the lowest similarity score. This is because this model is specifically designed to incorporate antonym relations where vectors of antonyms are designed by having negative similarity scores (numbers below zero).}
\end{itemize}
\item Finally, the performance metrics are computed over all the questions.
\end{enumerate}

\subsection{Results}
Table \ref{exp1} and Table \ref{exp2} include the results of Experiments 1 and 2 respectively.

 \begin{table}[h]
  \begin{center}
    \caption{Benchmark Experiment1}
    \label{exp1}
\scalebox{0.8}{
    \begin{tabular}{|c|ccc|ccc|ccc|}
 \hline
\multicolumn{1}{|c|}{Approach}&\multicolumn{3}{c|}{Dev 100} & \multicolumn{3}{c|}{Test 400}&\multicolumn{3}{c|}{All 500}\\
&Precision&Recall&Fscore&Precision&Recall&Fscore&Precision&Recall&Fscore\\
\hline
Polyglot (2013)&0.23&0.14&0.18&0.22&0.14&0.17&0.22&0.14&0.17\\
Altowayan (2016)&0.24&0.19&0.21&0.25&0.21&0.23&0.25&0.2&0.22\\
AraVec\_cbow (2017)&0.17&0.17&0.17&0.22&0.22&0.22&0.19&0.19&0.19\\
AraVec\_sg (2017)&0.18&0.18&0.18&0.19&0.19&0.19&0.19&0.19&0.19\\
fastText (2018)&0.1&0.1&0.1&0.15&0.15&0.15&0.14&0.14&0.14\\
ArWordVec\_SG3 (2020)&0.17&0.07&0.1&0.16&0.07&0.1&0.16&0.07&0.1\\
ArWordVec\_SG5 (2020)&0.12&0.05&0.07&0.15&0.07&0.09&0.14&0.06&0.09\\
AraBERT (2020)&0.27&0.27&0.27&0.25&0.25&0.25&0.25&0.25&0.25\\
ArSphere (2021) &\textbf{0.93}&\textbf{0.93}&\textbf{0.93}&\textbf{0.76}&\textbf{0.76}&\textbf{0.76}&\textbf{0.8}&\textbf{0.8}&\textbf{0.8}\\
 \hline
\end{tabular}
}
 \end{center}
\end{table}

 \begin{table}[h]
  \begin{center}
    \caption{Benchmark Experiment2}
    \label{exp2}
\scalebox{0.8}{
    \begin{tabular}{|c|ccc|ccc|ccc|}
 \hline
\multicolumn{1}{|c|}{Approach}&\multicolumn{3}{c|}{Dev 100} & \multicolumn{3}{c|}{Test 400}&\multicolumn{3}{c|}{All 500}\\
&Precision&Recall&Fscore&Precision&Recall&Fscore&Precision&Recall&Fscore\\
\hline
Polyglot (2013)&0.57&0.34&0.43&0.51&0.33&0.4&0.52&0.33&0.41\\
Altowayan (2016)&0.47&0.37&0.41&0.4&0.33&0.36&0.42&0.33&0.37\\
AraVec\_cbow (2017)&0.52&0.51&0.52&0.51&0.5&0.5&0.51&0.5&0.5\\
AraVec\_sg (2017)&0.59&0.58&0.59&0.56&0.56&0.56&0.57&0.56&0.56\\
fastText (2018)&0.63&0.62&0.62&0.59&0.58&0.59&0.6&0.59&0.59\\
ArWordVec\_SG3 (2020)&0.64&0.27&0.38&0.6&0.27&0.37&0.61&0.27&0.38\\
ArWordVec\_SG5 (2020)&0.57&0.24&0.34&0.61&0.28&0.38&0.6&0.27&0.37\\
AraBERT (2020)&0.49&0.49&0.49&0.45&0.45&0.45&0.46&0.46&0.46\\
ArSphere (2021) &\textbf{0.93}&\textbf{0.93}&\textbf{0.93}&\textbf{0.76}&\textbf{0.76}&\textbf{0.76}&\textbf{0.8}&\textbf{0.8}&\textbf{0.8}\\
 \hline
\end{tabular}
}
 \end{center}
\end{table}
{
\subsection{Comments on the Results}
All the models performed better in Experiment 2. This is because distributional models tend to place vectors of words that occur in the same context in close proximity (having high similarity scores) which is mostly the case for antonyms. It can be shown that the best performing model is ArSphere which is specifically trained to detect antonymy. The other models are not specifically trained to detect antonymy and they perform well in other relations such as detecting synonymy. From the results, it is also shown that the second best performing model is fastText \cite{Grave} followed by AraVec SG model \cite{Soliman}. }

\section{Conclusion and Future Work}
We have established the first closest opposite questions dataset for the Arabic language: ``ARCOQ". The dataset is beneficial for antonymy detection specifically and question answering in general. Moreover, a benchmark is developed comparing the major existing Arabic embedding models on ARCOQ. The benchmark shows that there still exists room for improving Arabic word embeddings on the aspect of antonymy. In the future, the presented benchmark can be built upon including other models either than embedding models. 

\bibliographystyle{unsrt}      % basic style, author-year citations
\bibliography{Ref}  % name your BibTeX data base

\end{document}